\documentclass[letterpaper, 10 pt, conference]{ieeeconf}  

\IEEEoverridecommandlockouts                              

\let\labelindent\relax
\usepackage[inline]{enumitem}
\usepackage{graphics} 
\usepackage{graphicx}
\usepackage{epsfig} 
\usepackage{mathptmx} 
\usepackage{times} 
\usepackage{amsmath} 
\usepackage{amssymb}  
\usepackage{mathrsfs}
\graphicspath{{./img/}}
\usepackage[table,xcdraw]{xcolor}
\usepackage{booktabs}
\usepackage{multirow}
\usepackage{url}
\usepackage{subcaption}
\usepackage{hyperref}

\DeclareMathOperator*{\argmin}{arg\,min}
\DeclareMathOperator{\st}{s.t.}

\title{\LARGE \bf
Driving from Vision through Differentiable Optimal Control
}
\author{Flavia Sofia Acerbo$^{1,2,3}$,
	Jan Swevers$^{2,4}$,
	Tinne Tuytelaars$^{3}$ and
	Tong Duy Son$^{1}$%
\thanks{$^{1}$Siemens Digital Industries Software, Leuven, Belgium}%
\thanks{$^{2}$Dept. of Mechanical Engineering, KU Leuven, Belgium}%
\thanks{$^{3}$Dept. of Electrical Engineering (ESAT), KU Leuven, Belgium}%
\thanks{$^{4}$Flanders Make@KU Leuven, Leuven, Belgium}%
\thanks{Corresponding author email: {\tt flavia.acerbo@siemens.com}}%
\thanks{\tiny{© 2024 IEEE. Personal use of this material is permitted. Permission from IEEE must be obtained for all other uses, in any current or future media, including reprinting/republishing this material for advertising or promotional purposes, creating new collective works, for resale or redistribution to servers or lists, or reuse of any copyrighted component of this work in other works.}}%
}

\begin{document}

\maketitle
\thispagestyle{empty}
\pagestyle{empty}

\begin{abstract}
This paper proposes DriViDOC: a framework for Driving from Vision through Differentiable Optimal Control, and its application to learn autonomous driving controllers from human demonstrations. 
DriViDOC combines the automatic inference of relevant features from camera frames with the properties of nonlinear model predictive control (NMPC), such as constraint satisfaction.
Our approach leverages the differentiability of parametric NMPC, allowing for end-to-end learning of the driving model from images to control. The model is trained on an offline dataset comprising various human demonstrations collected on a motion-base driving simulator. During online testing, the model demonstrates successful imitation of different driving styles, and the interpreted NMPC parameters provide insights into the achievement of specific driving behaviors. Our experimental results show that DriViDOC outperforms other methods involving NMPC and neural networks, exhibiting an average improvement of 20\% in imitation scores. 

\end{abstract}

\section{INTRODUCTION}
Mobile robotic systems, such as autonomous vehicles (AVs), rely on the integration of several key components, including perception, state estimation, planning, and control. This integration typically follows one of two paradigms: end-to-end or modular. In end-to-end architectures, all system components are jointly optimized towards the final control objective \cite{chen_end--end_2024,tampuu_survey_2022}. Imitation learning (IL) algorithms are central to these architectures, enabling the system to mimic demonstrated behaviors from high-dimensional sensor data, such as RGB images. Through this process, the model automatically identifies which visual patterns are necessary to accomplish the demonstrated driving tasks. However, these solutions are predominantly based on neural networks, as the overall policy must be differentiable. This black-box nature raises concerns about safety, interpretability and sample efficiency of data, particularly given the limited availability of human demonstrations. 
On the other hand, modular frameworks separate the system components, allowing for the use of model-based approaches, especially in planning and control. Nonlinear model predictive control (NMPC) is a prime example of a successful model-based technique in this context \cite{batkovic_experimental_2023, allamaa_real-time_2024}. NMPC makes decisions by optimizing the future states of the system over a time horizon, integrates safety constraints, and provides more interpretable behavior due to its reliance on a model. However, NMPC may struggle with the flexibility required to imitate a wide range of human driving behaviors and demands meticulous, time-consuming tuning of its cost function.

\begin{figure}[t]
    \centering
    \def\svgwidth{\columnwidth}
    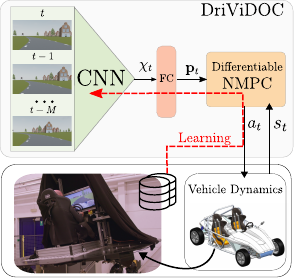
    \caption{DriViDOC architecture: at time $t$, the visually encoded information by the neural network $\chi_t$ serves to compute the parameters $\mathbf{p}_t$ of an NMPC, which controls a high-fidelity vehicle dynamics model used for human-in-the-loop testing on a hexapod platform. The driving model is learned end-to-end from pixels to control, based on human demonstrations collected offline on the platform.}
    \label{fig:archnnmpc}
\end{figure}

This paper proposes a framework that combines the strengths of both approaches, enabling safer and more interpretable imitation learning while also providing an online parameter tuning mechanism for NMPC cost functions. 
By exploiting the \emph{differentiability} of NMPC, we can compute the sensitivity of the optimal control actions with respect to the NMPC parameters. This allows NMPC to be integrated as a differentiable layer within a learning-based policy, with its parameters dynamically adjusted by preceding layers. As a result, the framework can map high-dimensional data to safe and feasible low-level control actions, all without disrupting the end-to-end pipeline, as illustrated in Fig.~\ref{fig:archnnmpc}.

The contributions of this paper can be summarized as:
\begin{enumerate}
    \item development of DriViDOC (Driving from Vision through Differentiable Optimal Control), an end-to-end differentiable driving model cascading a deep convolutional network with NMPC, mapping raw camera images to control actions through dynamic NMPC parameters,
    \item behavioral cloning on an offline dataset of different human driving styles, collected on a hexapod driving simulator, 
    \item evaluation of DriViDOC performance, in closed-loop simulations, compared to other state-of-the-art methods involving both neural networks and NMPC, showing the advantages of considering the NMPC already in the training phase and the value of learning latent representations from camera images.
\end{enumerate}
The paper is structured as follows: first, we discuss related work on the integration of IL and NMPC; second, we describe the main components of our methodology; third, we provide implementation details for learning the driving behaviors of various human drivers; finally, we present closed-loop simulation testing results of the learning and compare our method with alternative approaches.

\section{RELATED WORK}
Several methods have been developed to integrate end-to-end autonomous driving controllers with MPC. Typically, MPC acts as a low-level tracker for trajectories generated by learning-based planners, enhancing safety through its constraint-satisfaction capabilities \cite{tearle_predictive_2021, reiter_hierarchical_2023}. However, these approaches generally do not incorporate MPC into the learnable policy itself.

Alternatively, MPC, and more broadly optimal control, can serve as a substitute for neural network policy structures in learning-based algorithms. One approach involves tuning optimal control cost functions from demonstrations using inverse optimal control techniques, which align the features of demonstrated trajectories with the planned ones \cite{kuderer_learning_2015}. Another strategy, enabled by differentiable optimal control, involves directly learning MPC parameters through IL or reinforcement learning, as demonstrated in \cite{amos_differentiable_2018, gros_data-driven_2020}. However, in these studies, MPC is used as the entire policy without integrating other learning components.

Recent literature has also explored combining differentiable MPC with neural networks in larger end-to-end architectures. A notable example is \cite{karkus_diffstack_2023}, where differentiable MPC is integrated into a complete autonomous driving pipeline. Their experiments with real-world datasets show potential advantages over modular frameworks with non-differentiable components. However, the MPC parameters remain static, which the authors recognize as a limitation in fully capturing human driving behavior. Another study \cite{romero_actor-critic_2024} investigates the dynamic adaptation of MPC parameters for reinforcement learning on drones, using a fully connected network from low-dimensional observations. The use of visual data is shown by \cite{xiao_learning_2023}, which utilizes a differentiable MPC with a dynamically learned cost based on the current occupancy map for robotic navigation in human-occupied spaces.

Finally, our recent research investigates the application of MPC and IL to achieve human-like autonomous driving \cite{acerbo_evaluation_2023}. This study evaluates the effectiveness of tuning differentiable MPC based on demonstrations of human steering behavior and subsequently introduces a closed-loop learning algorithm that leverages MPC differentiability to jointly train a neural network setpoint generator. However, the proposed framework primarily relies on static MPC parameters, with the only dynamic elements being the setpoints generated by the fully connected network, from predetermined feature inputs.
In contrast, DriViDOC enables end-to-end learning from images, and allows for the dynamic parameterization of many MPC cost function parameters, including weights. 
Moreover, we compare this approach with others that integrate both MPC and neural networks, conducting more extensive experiments that involve a diverse range of human driving demonstrations.

\section{METHODOLOGY}

In our approach, an NMPC incorporates various parameters in its objective function, wherein their relative values directly influence the control style (e.g. tradeoff between desired speed and applied acceleration). Traditionally, these parameters are manually set, but with the differentiability of the NMPC, we can automatically learn them. To capture the intricacies of multiple human driving behaviors, DriViDOC uses \emph{dynamic parameterization}, i.e. instead of assuming a fixed set of NMPC parameters, different combinations are generated during execution based on the driving context. This driving context is embedded from front camera images by a convolutional neural network (CNN), which is trained to predict the NMPC parameters that best replicate the behavior of a given human driver.
In the following we introduce the three key components of DriViDOC: the NMPC formulation, the calculation of its derivatives with respect to parameters, and the dynamic parameterization given by its connection with the neural network layers.

Consider a generic \emph{parametric} NMPC, which at time $t$ computes the optimal action $a_t \in \mathbb{R}^{n_u}$ for the current state $s_t \in \mathbb{R}^{n_x}$. It does so by optimizing for future states $\mathbf{x} \in \mathbb{R}^{N n_x}$ and controls $\mathbf{u} \in \mathbb{R}^{(N-1) n_u}$. According to the receding horizon principle, the optimal action to be applied to the system is $a_t = u_0$. The scalar objective function $l: \mathbb{R}^{N n_x} \times \mathbb{R}^{(N-1) n_u} \times \mathbb{R}^{n_p} \xrightarrow{} \mathbb{R}$, composed of a stage cost $l_k$ and an optional terminal cost $l_N$, depends on a set of parameters $\mathbf{p} \in \mathbb{R}^{n_p}$. This corresponds to solving the following nonlinear program (NLP):
\begin{equation}\label{eq:mpcgeneric}
\begin{aligned}
    \min_{\mathbf{x},\mathbf{u}}  \quad & \sum_{k=0}^{N-1} l_k(x_k, u_k, \mathbf{p}) + l_N(x_N, \mathbf{p})& \\
    \st \quad  & x_0 = s_t, &\\
    & x_{k+1} = f(x_k, u_k), &\forall k = 0,...,N-1, \\
    & h(x_k, u_k) \leq 0, &\forall k = 0,...,N-1. \\
\end{aligned}
\end{equation}
We assume that the model $f$ and the inequality constraints $h$ do not depend on $\mathbf{p}$; in other words, vehicle dynamics, as well as state and actuator constraints, are known and fixed. Additionally, we assume that the current state $s_t$ can be measured or estimated using other sensors.
We also adopt a \emph{multiple shooting} solution strategy for (\ref{eq:mpcgeneric}). In this approach, the optimization variables are the both discretized states $\mathbf{x}$ and controls $\mathbf{u}$. These variables are computed by integrating $f$, which represents a discretized version of the model dynamics characterized by ordinary differential equations, across a prediction horizon consisting of $N$ equally spaced steps. The solution to the NLP can be obtained using any numerical solver of choice.

To learn the parameters $\mathbf{p}$ through gradient-based optimization, we need to formulate a differentiable version of the NMPC. In this paragraph, we discuss how to compute the derivatives of a solution of the NLP with respect to its parameters. 
The Lagrangian of Eq.~(\ref{eq:mpcgeneric}) is:
\begin{equation}
	\begin{aligned}
		\mathscr{L}(\mathbf{x}, \mathbf{u},\mu, \lambda, \mathbf{p}) = \quad &\sum_{k=0}^{N-1} l_k(x_k, u_k, \mathbf{p}) + l_N(x_N, \mathbf{p}) \\ 
		+ & \mu^Th(\mathbf{x},\mathbf{u}) + \lambda_0^T (x_0 - s_t) \\
		+ &\sum_{k=0}^{N-1} \lambda^T_{k+1}\left[ f(x_k, u_k) - x_{k+1} \right],
	\end{aligned}
\end{equation}
where $\mathbf{\mu} \in \mathbb{R}^{N}, \mathbf{\lambda} \in \mathbb{R}^{N}$ are the multipliers associated to the dual problem.
Now let us assume that $l$, $f$ and $h$ are twice-differentiable and that $z^* = [\mathbf{x}^*, \mathbf{u}^*]$ is a local minimum of (\ref{eq:mpcgeneric}). If the linear independence constraint qualification (LICQ) condition holds at $z^*$, then we can find $\mu^*$ and $\lambda^*$ that satisfy the following Karush-Kuhn-Tucker (KKT) conditions:
\begin{align}
    \nabla_z \mathscr{L}(\mathbf{x}^*, \mathbf{u}^*,\mu^*, \lambda^*, \mathbf{p}) & = 0, \\
    x_0^* &= s_t, \\
    x_{k+1}^* - f(x_k^*, u_k^*) & = 0, \forall k = 0,...,N-1, \\
    h(x_k^*, u_k^*) &\leq 0, \forall k = 0,...,N-1, \\
    \mu_k^* &\geq 0, \forall k = 0,...,N-1, \\
    \mu_k^*h(x_k^*, u_k^*) &= 0, \forall k = 0,...,N-1.
\end{align}
From \cite{noauthor_chapter_1983}, we know that if, in addition to LICQ, also second order sufficient condition (SOSC) and strict complementarity hold for $z^*$, 
then $z^*$ is a \emph{local isolated} minimizing point of (\ref{eq:mpcgeneric}) and $\mu^*$ and $\lambda^*$ are unique. In this case, we can write the KKT conditions as an implicit function $F(z,\mathbf{p}) \iff z = G(\mathbf{p})$, where the matrix $F$ is guaranteed to be non-singular. The derivatives are given by the \emph{implicit function theorem} as:
\begin{equation}\label{eq:gradient}
        	\frac{\partial G_\mathbf{p}}{\partial \mathbf{p}} = -\left(\frac{\partial F}{\partial z}\right)^{-1}\frac{\partial F}{\partial \mathbf{p}}.
\end{equation}
In learning algorithms, it is usually not necessary to compute the full Jacobian of (\ref{eq:gradient}). Rather, it is needed to compute the \emph{adjoint sensitivities}, that are the vector-times-Jacobian product, given any vector $\bar{z}$. In backpropagation, $\bar{z}$ happens to be the gradient of a scalar loss function $L$, e.g. behavioral cloning (BC) imitation loss, with respect to $z$. Then, by the chain rule, the gradient of $L$ with respect to the NMPC parameters $\mathbf{p}$ is defined as:
\begin{equation}\label{eq:adj}
    \nabla_{\mathbf{p}} L = \left[ \frac{\partial G_\mathbf{p}}{\partial \mathbf{p}} \right]^T \nabla_{z} L.
\end{equation}
The computation of (\ref{eq:adj}) can be done in CasADi \cite{andersson_casadi_2019} at the cost of a linear system solve and a reverse mode algorithmic differentiation pass, as shown in \cite{andersson_sensitivity_2018}. 

In DriViDOC, the parameters $\mathbf{p}$ of the NMPC objective function are not constant, but varying in time and depending on visual data. More specifically, a CNN takes a series of frames $\{I_t, I_{t-1},...,I_{t-M}\}$ and processes them by a sequence of convolutional layers. These layers act as a visual encoder and the resulting latent representation $\chi_t$ is given to a sequence of fully-connected (FC) layers with $n_p$ heads, which predict the parameters $\mathbf{p}(\mathbf{\chi}(I_t, I_{t-1},...,I_{t-M}))$ for the NMPC. The last layers of the FC network should be designed to ensure the positive semi-definiteness of the learned cost.

To integrate the NMPC and the CNN layers, the software framework of DriViDOC interfaces CasADi and PyTorch, effectively replacing the automatic differentiation of PyTorch by the one of CasADi. Differently from other software frameworks for differentiable optimal control, e.g. \cite{amos_differentiable_2018,wang_pypose_2023,drgona_neuromancer_2023}, our implementation is not restricted on a specific solver for the NMPC (e.g. iLQR), but allows us to employ optimizers such as IPOPT \cite{wachter_implementation_2006}, which are better suited to manage nonlinear constraints (including those on states) and poor initial guesses. This approach ensures more reliable solutions for the NMPC, thereby enhancing the overall robustness of the differentiation process. 


\section{IMPLEMENTATION}
In this section we present the application of DriViDOC to the specific context of learning autonomous driving controllers from a diverse driving dataset collected on the hexapod platform shown in Fig. \ref{fig:archnnmpc}. First, we provide more details about the dataset; second, we describe the architecture of the CNN and the NMPC formulation; and finally, we outline the training procedure.
\subsection{Dataset}
The dataset was collected on a 6-degree of freedom Stewart platform (MOOG MB-EP-6DOF/24/2800kg) \cite{noauthor_moog_nodate}. The scenario, designed in Simcenter Prescan \cite{noauthor_simcenter_nodate}, consists of a one-lane road 4.5m wide, delimited with a dashed line at both sides and with randomly placed road furniture. The road presents 8 clothoidal curves.
Following the convention of positive curvature for left curves and negative for right one, the radii of curvature are the following, in meters: $[90, -90, 100, -100, 110, -110, 120, -120]$. The vehicle is modelled in Simcenter Amesim \cite{noauthor_simcenter_nodate-1} as a 15 DoF vehicle, based on real experimental data of a Siemens Simrod car \cite{debille_simrod_nodate}. The model takes into account chassis, tire, powertrain and brake dynamics, steering wheel torque feedback, and aerodynamic effects.
11 drivers participated in the study, and were asked to perform 2 or 3 trial laps before recording. Following the trial, each participant drove for 5 laps, corresponding to a duration of approximately 15 minutes. The drivers were asked to stay in the lane and keep a speed between 60 and 80 km/h. The drivers could monitor their current speed via a speedometer present on the screen.

\begin{figure}[tb]
    \begin{subfigure}{0.48\columnwidth}
        \includegraphics[width=\linewidth]{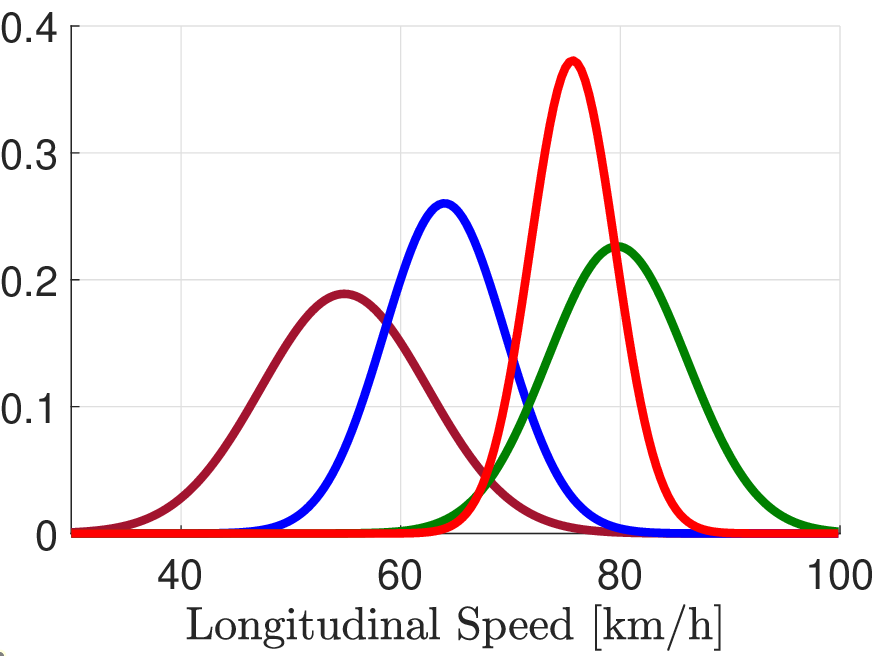}
        \caption{}
        \label{fig:distribvx}
    \end{subfigure}%
    \hfill
    \begin{subfigure}{0.48\columnwidth}
        \vspace{0.3cm}
        \def\svgwidth{\columnwidth}
        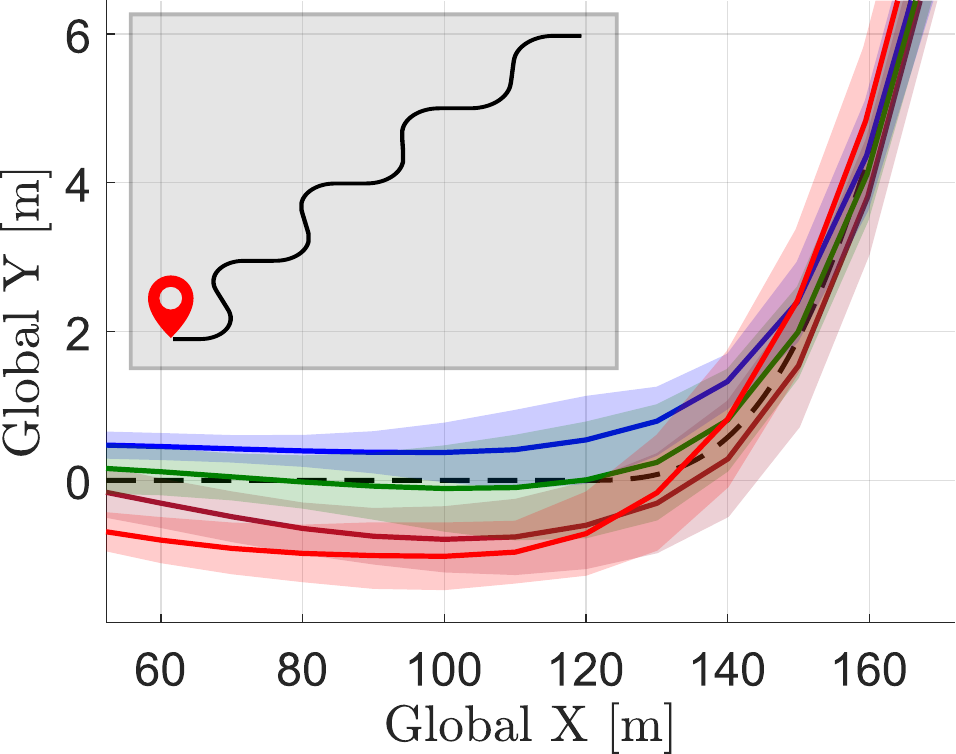
        \caption{}
        \label{fig:distribd}
    \end{subfigure}
    \caption{Illustrations of four different driving styles present in the dataset. In (a): the Gaussian probability distributions fitted to $v_x$ for four different drivers. In (b): position with respect to the centerline (dashed line) for the same drivers while entering the first curve of the track, in global cartesian coordinates. Inside the plot, a minimized map of the track is included.}
    \label{fig:driversdistr}
    
\end{figure}

The resulting dataset consists of 113,768 samples, with a frame rate of 10Hz. Each sample is composed of an image $I_t$ and the set of state variables computed from the vehicle model, encompassing linear and angular positions, velocities and accelerations. The data shows diverse behaviors between drivers around many aspects.
In Figure \ref{fig:distribvx}, we provide an illustrative example of the distribution of longitudinal speed for four sample drivers across the recorded laps, exhibiting variations in both mean value and variance. Furthermore, in Figure \ref{fig:distribd}, we present a map of the experimental track and depict the positions within the lanes for the same four sample drivers as they approach the first curve. This visualization highlights the different preferred approaches, including maintaining the center, staying on the inner side, or anticipating the curve by positioning on the outside and entering at different timings.


\subsection{NMPC and CNN Design}
The NMPC is designed to perform path following based on a dynamic single-track vehicle model, as described in \cite{allamaa_real-time_2022}, in a Frenet coordinate system with respect to the road centerline, as defined in \cite{acerbo_evaluation_2023}. The control actions $\mathbf{u}$ are the steering wheel angle rate $\dot{\delta}$ and the normalized throttle $t_r$. The states are the linear velocities $v_x, v_y$, in a car body frame with $x$ pointing forward, the yaw angle rate $\dot{\psi}$, the Frenet coordinates $\sigma, d, \theta$, and the steering wheel angle $\delta$. The Frenet coordinates evolve according to the road curvature $\kappa(\sigma)$, that we assume as known. 
The dynamic equations are discretized and integrated over the prediction horizon with Runge Kutta 4. The horizon is set at 1.5 seconds, with a sampling rate of 0.1 seconds, resulting in N = 15.
Following the notation of (\ref{eq:mpcgeneric}), we also define the objective function:
\begin{equation}\label{eq:objective}
\begin{aligned}
    l_k =&W_d(d_k - \bar{d})^2 + W_v(v_{xk} - \bar{v_x})^2
    + W_{\dot{\delta}}(\dot{\delta}_k)^2 + W_{t_r}(t_{rk})^2+\\
    + &\gamma(d_k^2 + \theta_k^2 + \delta_k^2+ \dot{\delta}_k^2 + t_{rk}^2),
\end{aligned}
\end{equation}
where $\mathbf{p} = [ W_d, \bar{d}, W_v, \bar{v_x}, W_{\dot{\delta}}, W_{t_r}]^T$ and the last term with $\gamma$ is added for regularization and to ensure SOSC. No terminal cost $l_N$ is defined. In (\ref{eq:objective}), before the variables in are squared and summed, they are all scaled to be between [-1,1]. The inequality constraints consist of a lane boundary constraint $-\frac{w}{2} \leq d_k \leq \frac{w}{2}$, a speed limit constraint $v_{min} \leq v_{xk} \leq v_{max}$, and the rest constraining all other states and control inputs to a feasible set, according to system specifications. The solution of the NLP is computed through IPOPT \cite{wachter_implementation_2006}. 

The CNN architecture is inspired by \cite{bojarski_end_2016}, with some modifications on the input channels and on the final decision layers. This choice is taken prioritizing a lightweight structure, needed for future real-time deployment of the model. More specifically, let $I_t$ denote the $t$-th frame of a front camera stream. Past information is included by concatenating the last $M = 3$ frames on the channels as input to our CNN, so that it accepts $3(M+1)$ channels. Then, the concatenated frames are processed by five 2x2 convolutional layers of 24, 36, 48, 64, 64 output channels each. The resulting 1152 element latent variable $\chi_t$ is fed to a two-layer fully-connected network with 100 and 50 neurons. Finally, $n_p = 6$ one-layer heads predict the parameters $\mathbf{p}_t$. The heads associated to the objective function weights have a ReLU activation function on the output, to guarantee positive semi-definiteness of the learned cost, while the ones associated with offsets have a $tanh$, to map the offsets between limits that are compatible with the constraints. We refer to the learnable weights and biases of the CNN + FC network as $\Theta = \{W_{k=1,..,5}^{(i)}, b_{k=1,..,5}^{(i)}, W_{\text{fc}}^{(i, j)}, b_{\text{fc}}^{(j)}\}$.
\subsection{Open-Loop Training}
The human demonstrations are composed by tuples of images, states and actions, denoted as $\{I^h_t, s^h_t, a^h_t\}$.
Firstly, image data is augmented via random perturbations of certain vehicle states, as this is shown to reduce the effect of covariate shift for BC \cite{bojarski_end_2016}. 
Secondly, the unperturbed data is split such that the samples from the curves with 110 m and -110 m radius of curvature are kept for testing and validation. The camera frames are first cropped and resized to 64x200 pixels and then normalized.
The data is then balanced by downsampling straight driving samples and upsampling curved driving ones. The actions $a^h_t$ are scaled between 0 and 1.

Finally, the driving model is trained with BC, on a training dataset of $D_{train}$ size, by solving the following supervised learning problem,
\begin{equation}\label{eq:bc}
    \Theta^* = \argmin_{\Theta} \sum_{t=1}^{D_{train}} ||a^h_{t} - a_t(I^h_t, s^h_t) ||_2^2,
\end{equation}
where $a^h_t$ is sampled from the demonstrations dataset, whereas $a_t$ represents the action computed by DriViDOC based on $\{I^h_t, s^h_t\}$, which are also sampled from the dataset together with $a^h_t$.
Problem (\ref{eq:bc}) is solved using the Adam Optimizer with an initial learning rate of $10^{-4}$.
A separate model is trained for each driver to assess whether the specific demonstrated driving patterns are accurately imitated. The training is performed in three phases. To start, the CNN is trained separately for 100 epochs on all samples, coming from all drivers, to predict $\delta$ and $v_x$ from camera frames. The optimal validation loss model is subsequently fine-tuned individually for each driver, utilizing 25 epochs of data specific to that driver. Then, the FC heads are modified and the NMPC is added as last layer of the driving model. CNN and NMPC together are finetuned for 25 epochs to predict $\dot{\delta}$ and $t_r$ from camera frames and states. We train with a batch size of 10 on an NVIDIA GeForce RTX 3060 GPU (for CNN tensors) and on a Ryzen ThreadRipper 3995WX Pro (for NMPC tensors). The pretraining of the CNN on all drivers takes approximately 33 hours, while the finetuning on one driver takes about 1 hour for the CNN alone and 40 hours for both the CNN and NMPC together.

\section{RESULTS}
\subsection{Closed-Loop Testing}
The learned driving models are tested in closed-loop simulations, with the same high-fidelity car model and scenario used for data collection on the hexapod platform. In the following section, we present the results related to these simulations.
\subsubsection{State Simulations}\label{simulationsparag}  
\begin{figure}[tb]
\vspace{0.2cm}
    \centering
    \includegraphics[width=\columnwidth]{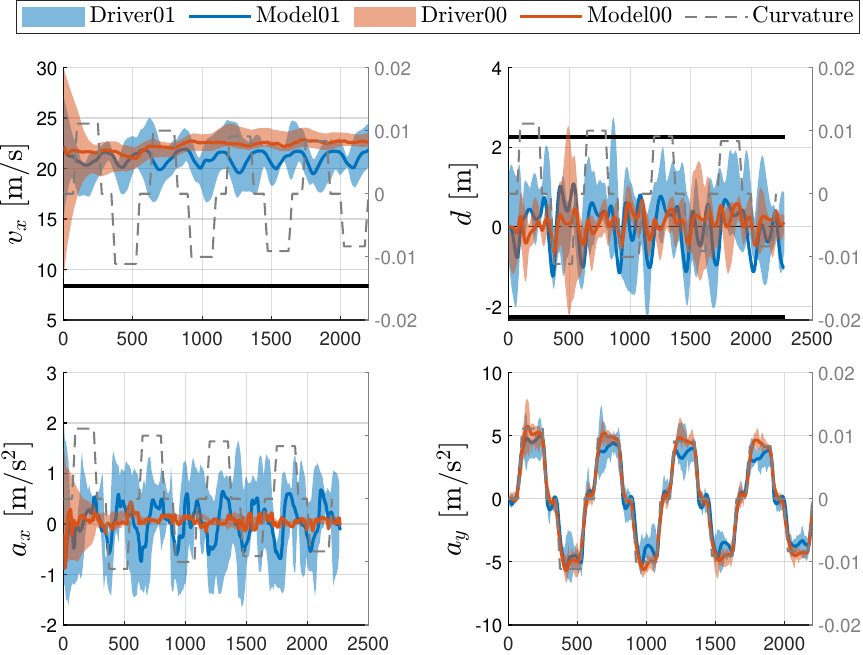}
    \caption{Relevant states from closed-loop simulations of two DriViDOC models, compared with the corresponding driver distribution. On the right axis, we indicate the curvature with a dashed line. In the $d$ plot, the lane boundaries are shown as horizontal solid lines.}
    \label{fig:simulations}
    
\end{figure}

The considered states for the evaluation are $v_x$, $d$, $a_x$ and $a_y$. In Fig. \ref{fig:simulations}, we show how these cues differ for two driving models, and how they qualitatively match the corresponding demonstrated driving style. To model the average driving style and its variance across the repeated laps, we do the following: given the collected states for one driver, we consider the mean $\mu_s(\sigma)$ and standard deviation $SD_s(\sigma))$ for each of them in every track point (with 1m tolerance). In Fig. \ref{fig:simulations} we plot $SD_s(\sigma), \forall \sigma = 0,...,L^{track}$. Driver00 exhibits constant speed behavior, reflected in the driving model learned from this data (Model00). Conversely, Driver01 displays oscillating speed patterns along curves, characterized by deceleration from the start to the mid-point and acceleration thereafter. Model01, trained on this data, captures both the speed oscillations and longitudinal acceleration trends resulting from throttle adjustments.

\subsubsection{NMPC Parameters} 
\begin{figure}[tb]
\vspace{0.2cm}
    \centering
    \includegraphics[width=\columnwidth]{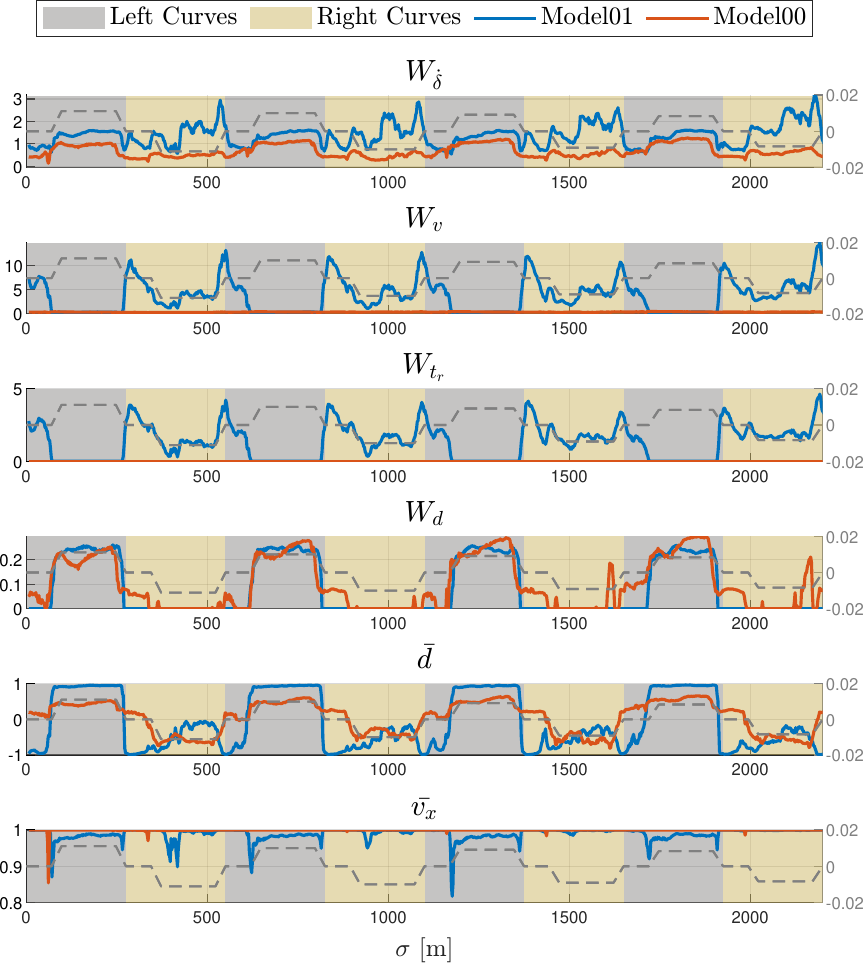}
    \caption{On the left axis of each plot, parameters $\mathbf{p}(t)$ values from closed-loop simulations of DriViDOC models trained for 2 different drivers. The offsets $\bar{d}$ and $\bar{v_x}$ are normalized between -1 and 1. On the right axis, we indicate the curvature with a dashed line.}
    \label{fig:params}
    
\end{figure}

Having a finite set of NMPC parameters increases the interpretability of DriViDOC since the difference in driving models can be attributed to the dynamic variations in the parameters of the NMPC objective function, shown in Fig. \ref{fig:params}. From that, we discover that the same driving model exhibits a different behavior in left and right curves, which is not evident in the state simulations. In right curves, the oscillating speed of Model01 is attributed to a decreasing and increasing behavior of $W_v$ and $W_{t_r}$. In left curves, Model01 exhibits instead a constant lower $W_v$ and null $W_{t_r}$. As a result, the vehicle decelerates until the longitudinal error ($|v_{x} - \bar{v_x}|_2^2$) becomes sufficiently large to trigger the control to accelerate again. Model00 shows instead a constant $W_v$, resulting in the vehicle maintaining a constant speed. 
Model00 shows also a preview behavior, anticipating the curve ahead when on the straight driving part, as it can be seen from the peaks of $W_d$ when curvature is zero. This aims at positioning the vehicle ahead of the curve and hence reducing $d$ variations when entering. From this analysis we note that the offsets are not meaningful setpoints per se, but that, combined with the weights, constitute a good parameter set for the NMPC to mimic the driving actions.

\subsubsection{Human-Likeness}
To quantify the performance of the driving models, we consider the demonstrations as a series of Gaussians as described in \ref{simulationsparag}, and we compute for each relevant state $s$ the following metrics:
\begin{itemize*}
    \item Absolute Error (AE) $= |s(\sigma) - \mu_s(\sigma)|, \forall \sigma = 0,...,L^{track}$,
    \item Z-score $ = |\frac{s(\sigma) - \mu_s(\sigma)}{SD_s(\sigma)}|, \forall \sigma = 0,...,L^{track}$, to take into account also the variance present in the demonstrations. Values that are too far from the driver distribution show a Z-score $>$ 3.
\end{itemize*}
We provide mean and standard deviation for both metrics and all drivers in Table \ref{tab:alldrivers}. On average, the errors are according to expectations, with some exceptions, due to shortcomings in the NMPC control actions.
Model02 exhibits $d$ values that deviate significantly from the demonstration distribution, primarily attributed to some rapid full throttle releases by the driver. Model03 shows a similar discrepancy on both $d$ and $a_x$ that can be traced back to an even more consistent use of full throttle release by the driver, resembling a discrete behavior and with zero standard deviation. Additionally, Model04 shows $a_x$ values slightly distant from the distribution, attributed to frequent use of the brake by the driver. In contrast, the NMPC, which inherently provides a more continuous control, consistently avoids fully releasing the throttle and does not control the brake. Consequently, it maintains a speed slightly higher than the mean. This leads to the learned steering action failing to accurately position the vehicle within the lane and to the simulated longitudinal acceleration diverging from the demonstrations during hard deceleration peaks. However, in this way, the NMPC avoids providing uncomfortable longitudinal acceleration and jerk to the passenger.
\begin{table*}[tb]
\vspace{0.1cm}
	\centering
	\caption{Performance of DriViDOC for each driver, averaged across the track length. Less optimal results (Z-score $>$ 2) are highlighted with a red colour.}
	\label{tab:alldrivers}
	\begin{tabular}{cllllllll}
		\hline
		\multicolumn{1}{l}{} & \multicolumn{2}{c}{$d$}                                    & \multicolumn{2}{c}{$v_x$}                                    & \multicolumn{2}{c}{$a_x$}                                      & \multicolumn{2}{c}{$a_y$}                                      \\ \hline
		\multicolumn{1}{l}{} & \multicolumn{1}{c}{AE $[m]$} & \multicolumn{1}{c}{Z-score} & \multicolumn{1}{c}{AE $[m/s]$} & \multicolumn{1}{c}{Z-score} & \multicolumn{1}{c}{AE $[m/s^2]$} & \multicolumn{1}{c}{Z-score} & \multicolumn{1}{c}{AE $[m/s^2]$} & \multicolumn{1}{c}{Z-score} \\ \hline
		Model00 & $0.18 \pm 0.14$ & $0.84 \pm 0.79$ & $0.36 \pm 0.29$ & $0.63 \pm 0.37$ & $0.08 \pm 0.11$ & $1.68 \pm 1.65$ & $0.20 \pm 0.17$ & $0.65 \pm 0.55$ \\ \hline
		Model01 & $0.43 \pm 0.32$ & $1.16 \pm 1.07$ & $0.49 \pm 0.38$ & $0.52 \pm 0.39$ & $0.22 \pm 0.18$ & $0.73 \pm 0.83$ & $0.37 \pm 0.29$ & $0.84 \pm 0.73$ \\ \hline
		Model02 & $0.51 \pm 0.32$ & $\textcolor{red}{3.39 \pm 2.21}$ & $1.69 \pm 0.86$ & $0.83 \pm 0.46$ & $0.16 \pm 0.14$ & $0.65 \pm 0.63$ & $0.53 \pm 0.40$ & $1.36 \pm 1.22$ \\ \hline
		Model03 & $0.61 \pm 0.37$ & $\textcolor{red}{2.51 \pm 2.38}$ & $1.37 \pm 0.69$ & $1.50 \pm 1.55$ & $0.25 \pm 0.25$ & $\textcolor{red}{2.46 \pm 4.42}$ & $0.56 \pm 0.31$ & $1.35 \pm 0.84$ \\ \hline
		Model04 & $0.42 \pm 0.32$ & $1.66 \pm 1.32$ & $1.21 \pm 0.77$ & $1.29 \pm 0.88$ & $0.51 \pm 0.47$ & $\textcolor{red}{2.01 \pm 2.00}$ & $0.46 \pm 0.33$ & $1.21 \pm 0.86$ \\ \hline
		Model05 & $0.35 \pm 0.26$ & $2.00 \pm 1.87$ & $0.88 \pm 0.74$ & $0.69 \pm 0.43$ & $0.29 \pm 0.24$ & $0.85 \pm 0.79$ & $0.29 \pm 0.21$ & $1.30 \pm 1.39$ \\ \hline
		Model06 & $0.44 \pm 0.32$ & $1.34 \pm 1.27$ & $1.40 \pm 1.18$ & $0.81 \pm 0.87$ & $0.23 \pm 0.15$ & $0.98 \pm 0.85$ & $0.43 \pm 0.32$ & $1.04 \pm 0.93$ \\ \hline
		Model07 & $0.27 \pm 0.19$ & $1.09 \pm 0.93$ & $0.62 \pm 0.52$ & $0.45 \pm 0.42$ & $0.22 \pm 0.20$ & $0.96 \pm 1.04$ & $0.29 \pm 0.24$ & $0.77 \pm 0.77$ \\ \hline
		Model08 & $0.41 \pm 0.26$ & $0.94 \pm 0.61$ & $1.00 \pm 0.71$ & $0.98 \pm 0.75$ & $0.19 \pm 0.17$ & $0.83 \pm 0.87$ & $0.41 \pm 0.34$ & $0.92 \pm 0.86$ \\ \hline
		Model09 & $0.41 \pm 0.27$ & $1.42 \pm 1.32$ & $0.53 \pm 0.51$ & $0.99 \pm 1.36$ & $0.13 \pm 0.12$ & $1.03 \pm 0.95$ & $0.31 \pm 0.29$ & $1.00 \pm 1.10$ \\ \hline
		Model10 & $0.17 \pm 0.12$ & $0.98 \pm 0.86$ & $1.42 \pm 0.87$ & $1.18 \pm 1.12$ & $0.25 \pm 0.23$ & $0.82 \pm 0.78$ & $0.41 \pm 0.30$ & $0.96 \pm 0.82$ \\ \hline
	\end{tabular}
\end{table*}

\subsection{Benchmarking}
In this section, we discuss experimental results of the benchmarking of DriViDOC against other baseline approaches involving neural networks and NMPC.
The benchmarking results are summarized in Table \ref{tab:baselines}. For each baseline approach, we report the mean absolute error (MAE) and the mean Z-score (MZ) over the track, with mean and standard deviation for all drivers. On average, DriViDOC shows better performance on all metrics, with an average 20\% relative improvement. In the following, we discuss the results for each baseline approach.
\begin{table}[tb]
\caption{Benchmarking studies. We report performance in terms of Mean Absolute Error (MAE) and Mean Z-score (MZ) over the track. Results are averaged over all drivers, except for the NOIMG baseline for which drivers 00,03,10 and 11 have been excluded (see text). Best results are highlighted in bold.}
\label{tab:baselines}
\resizebox{\columnwidth}{!}{
\begin{tabular}{ccccc
>{\columncolor[HTML]{C0C0C0}}c }
\hline
                        &     & TRACK           & SF              & NOIMG                    & DriViDOC (Ours)          \\ \hline
                        & MAE & $0.38 \pm 0.15$ & $0.59 \pm 0.15$ & $\mathbf{0.29 \pm 0.12}$ & $0.38 \pm 0.13$          \\ \cline{2-6} 
\multirow{-2}{*}{$d$}   & MZ  & $1.49 \pm 0.53$ & $2.51 \pm 0.83$ & $\mathbf{1.40 \pm 0.73}$ & $1.58 \pm 0.78$          \\ \hline
                        & MAE & $1.30 \pm 0.47$ & $2.62 \pm 2.09$ & $2.53 \pm 0.98$          & $\mathbf{1.00 \pm 0.45}$ \\ \cline{2-6} 
\multirow{-2}{*}{$v_x$} & MZ  & $1.25 \pm 0.43$ & $2.36 \pm 1.93$ & $1.81 \pm 0.28$          & $\mathbf{0.90 \pm 0.33}$ \\ \hline
                        & MAE & $0.45 \pm 0.15$ & $0.25 \pm 0.12$ & $0.36 \pm 0.12$          & $\mathbf{0.23 \pm 0.11}$ \\ \cline{2-6} 
\multirow{-2}{*}{$a_x$} & MZ  & $2.72 \pm 1.17$ & $1.20 \pm 0.74$ & $2.05 \pm 1.53$          & $\mathbf{1.18 \pm 0.59}$ \\ \hline
                        & MAE & $0.50 \pm 0.17$ & $0.67 \pm 0.39$ & $0.65 \pm 0.19$          & $\mathbf{0.39 \pm 0.11}$ \\ \cline{2-6} 
\multirow{-2}{*}{$a_y$} & MZ  & $1.35 \pm 0.38$ & $1.60 \pm 0.79$ & $1.51 \pm 0.24$          & $\mathbf{1.04 \pm 0.24}$ \\ \hline
\end{tabular}
}
\end{table}

\subsubsection{NMPC Tracker for a CNN Planner (TRACK)}
In this baseline, the CNN predicts $\bar{d}$ and $\bar{v_x}$ values, following a similar approach to that used in \cite{acerbo_evaluation_2023}. These predictions are learned in a manner such that, for each time $t$, the CNN maps the input frames sequence $(I_t, I_{t-1},...,I_{t-M}), M = 3$ to the corresponding output vector $[d_{t+T_f}, v_{x(t+T_f)}]^T$, where $T_f$ represents the NMPC prediction horizon (set to 1.5 seconds, corresponding to $N$ = 15). These predicted values then serve as setpoints in the NMPC's objective function as follows:
\begin{equation}\label{eq:mpctrack}
   \min_{\mathbf{x},\mathbf{u}}  \quad \sum_{k=0}^{N-1}(d_k - \bar{d})^2 + (v_{xk} - \bar{v_x})^2 + \gamma(\dot{\delta}_k^2 + \theta_k^2 + d_k^2 + t_{rk}^2).
\end{equation}
The CNN for these models is finetuned on data coming from the corresponding driver for 50 epochs, starting from the model trained on all drivers predicting speed and steering angle.

\begin{figure}[tb]\vspace{0.2cm}
    \centering
    \includegraphics[width=\columnwidth]{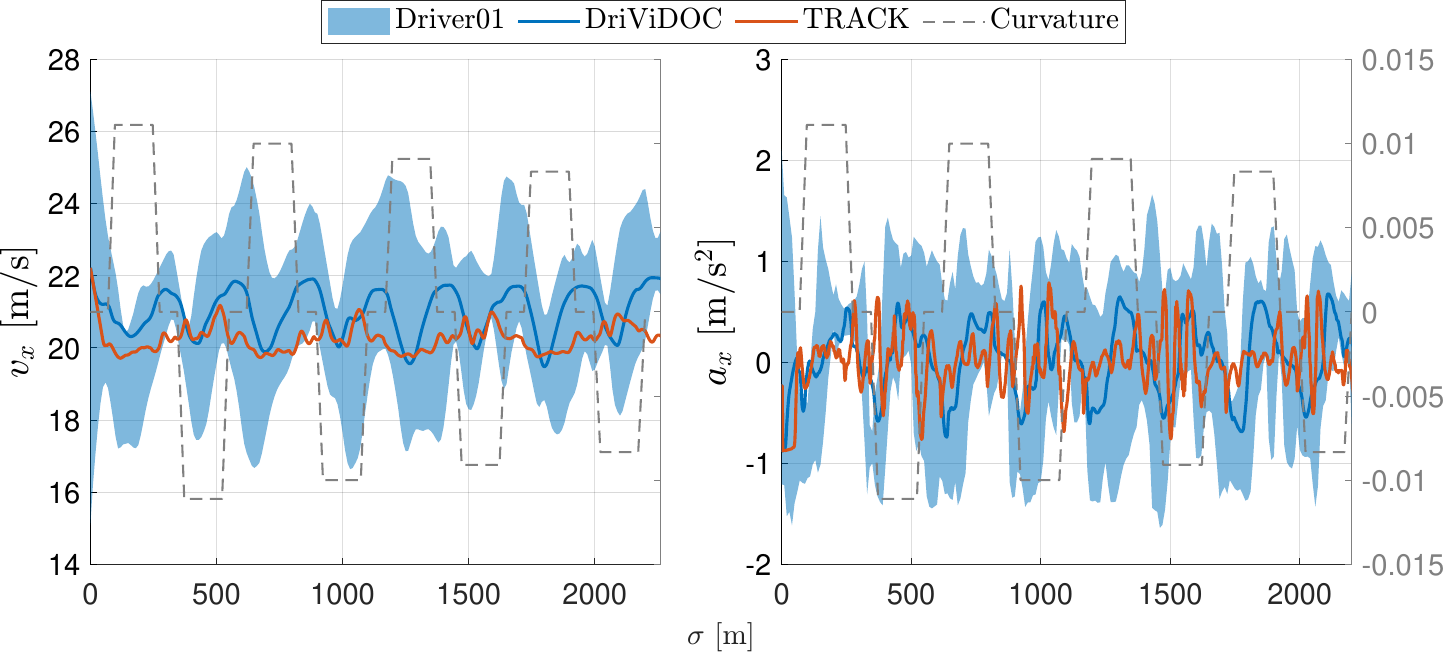}
    \caption{DriViDOC vs TRACK baseline for Driver01 ($v_x$ and $a_x$).}
    \label{fig:track}
\end{figure}

TRACK exhibits most of its problems in the longitudinal behavior, while having better match with the driver distribution on $d$, with respect to DriViDOC. Since TRACK does not learn the low level control actions, the NMPC provides an unnatural behavior on the throttle. This reflects on $a_x$ significantly far from the demonstrations, with a relative increase of 95\% in MAE compared to DriViDOC. Moreover, for $v_x$, the model sometimes tends to provide constant references, even though the trend is not. This can be attributed to the network predicting a setpoint only 1.5 seconds ahead. In situations where changes of $v_x$ are not significantly fast, the future state may be very close to the current one. As a result, when $v_{xt}$ is already within the distribution, the CNN predicts $\bar{v} \approx v_{xt}$. These phenomena can be both seen in Fig. \ref{fig:track}. Conversely, DriViDOC is less affected by a short prediction horizon, as the CNN adapts the parameters accordingly by taking into account the NMPC already in the training phase.

\subsubsection{NMPC as Safety Filter for the CNN Controller (SF)}
This method is adapted from \cite{tearle_predictive_2021}. In this approach, the CNN predicts steering angle and throttle as $a_{CNN} = [\delta, t_r]^T$. These actions are subsequently input into an NMPC to assess whether a trajectory to a terminal safe set of choice can be planned once they are applied. If not, the NMPC calculates the closest control action to ensure the feasibility of planning a trajectory to the safe set. Specifically, the NMPC is formulated as follows:
\begin{equation}\label{eq:mpcsf}
\begin{aligned}
   \min_{\mathbf{x},\mathbf{u}}  \quad & ||u_0 - a_{CNN}||_2^2 + \gamma\sum_{k=0}^{N-1}(\delta_k^2 + \theta_k^2 + d_k^2 + t_{rk}^2)& \\
    \st \quad &d_N = \theta_N = 0. &
\end{aligned}
\end{equation}
The terminal safe set we consider is when the vehicle is fully aligned with the road. 
The CNN for these models is finetuned as in the TRACK baseline.

The results with SF exhibit a clear trend toward worse imitation scores for $d$ and $v_x$, with a relative increase in MAE of 54\% and 162\%, respectively, compared to DriViDOC. Nonetheless, some small improvements relative to DriViDOC are noticeable, particularly for $a_x$ in 04 and 05 (drivers with throttle release and braking), as well as for 00 and 09 (drivers with quasi-constant throttle), with a mean relative reduction of 25\% in MAE with respect to DriViDOC. This occurs because, during training, the NMPC does not dictate the shape of the control actions and the CNN alone finds it easier to generate such actions. On average, however, SF appears more susceptible to the covariate shift problem, tending to deviate from the driver speed distribution. It is only when the vehicle approaches a state that will lead to constraint violation that the safety filter intervenes, correcting to the closest safe control actions. However, given that the vehicle already moves beyond the training distribution, errors accumulate, and the safety filter takes over more often. Consequently, SF frequently remains close to the constraint boundary without reverting to the distribution. In DriViDOC, instead, the NMPC intrinsically tends to keep the model in the driver distribution. We give an illustrative example of the difference between the two behaviors for $v_x$ in Fig. \ref{fig:sfvx}.

\subsubsection{Differentiable NMPC without Images (NOIMG)}
For this baseline, we aim to assess the impact of learning directly from raw representation, as opposed to using predefined features, which was demonstrated in \cite{acerbo_evaluation_2023}. To achieve this, we eliminate the convolutional layers. The handcrafted features, fed directly into the FC layers, are the following: $\chi_t = [v_{xt}, d_t, \theta_t, \kappa(\sigma_t), \kappa(\sigma_t+5), \kappa(\sigma_t+10), ..., \kappa(\sigma_t + 30)]^T$, i.e. a subset of the current state and samples of the road curvature, spanning from the current track point $\sigma_t$ to 30 meters ahead ($\sigma_t + 30$). The NMPC remains invariant with respect to DriViDOC. The model is trained separately for each individual driver using 25 epochs and starting from scratch. The training takes approximately 20 hours, which is less than DriViDOC, owing to the reduced size of the network and the absence of data augmentation.

\begin{figure}[tb]
\vspace{0.2cm}
    \begin{subfigure}{0.49\columnwidth}
        \includegraphics[width=\linewidth]{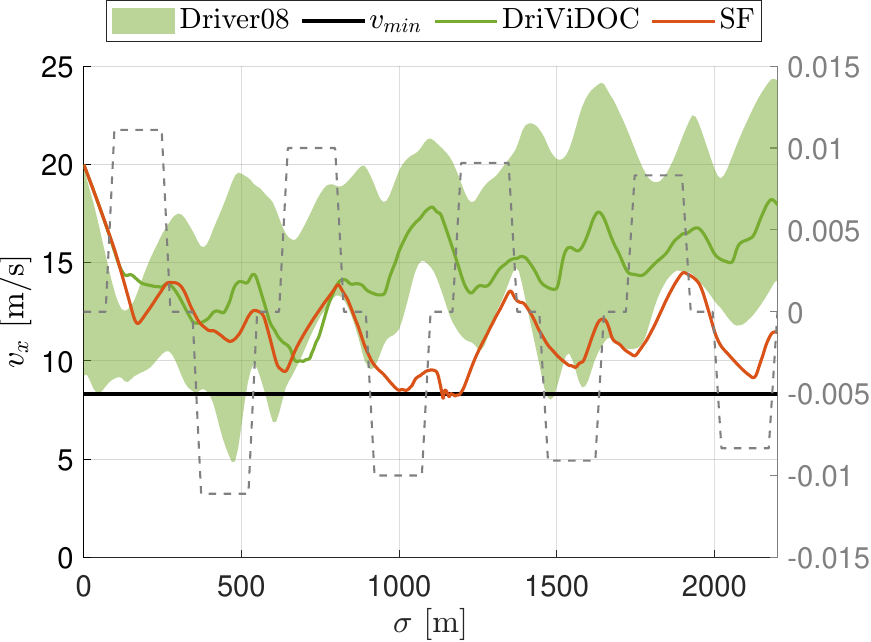}
        \caption{DriViDOC vs SF baseline for Driver06 ($v_x$).}
        \label{fig:sfvx}
    \end{subfigure}%
    \hfill
    \begin{subfigure}{0.49\columnwidth}
        \includegraphics[width=\linewidth]{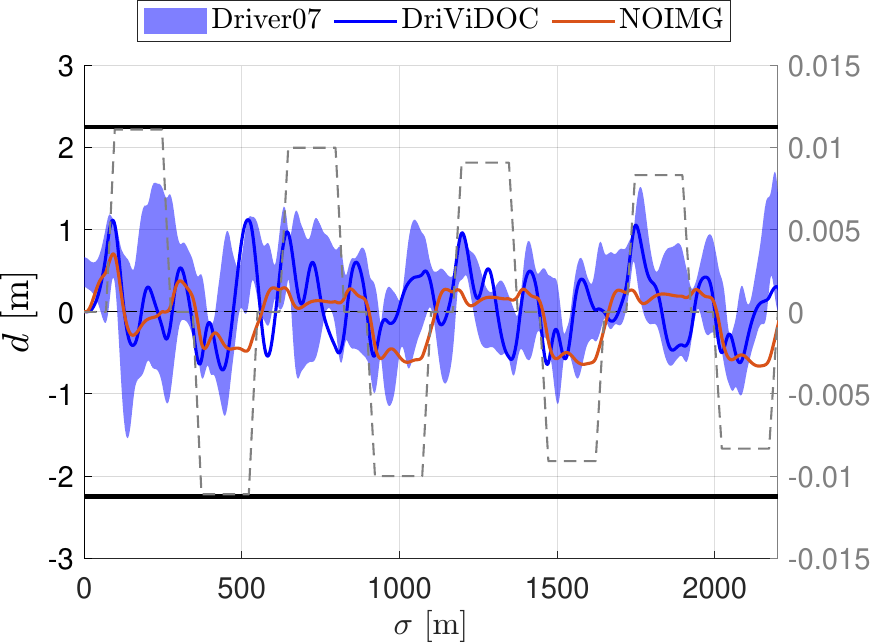}
        \caption{DriViDOC vs NOIMG baseline for Driver07 ($d$).}
        \label{fig:trackd}
    \end{subfigure}
    \caption{}
    \label{fig:combined}
\end{figure}

NOIMG exhibits inconsistent behavior across drivers. Specifically, for models 00, 01, 08, and 09, the BC training loss still does not converge after 25 epochs. This corresponds to an unacceptable closed-loop behavior, characterized by an immediate deceleration followed by a constant speed at the minimum value. We argue that models experiencing training convergence issues are those corresponding to drivers with higher average speeds.
For an easier comparison with DriViDOC and the other baselines, the NOIMG models exhibiting this behavior have been excluded from the average metrics calculation. However, even for the remaining models, there is a decrease in imitation scores, with the exception of $d$. 
In many cases, NOIMG struggles to match the changing dynamics of certain states, and provides instead a more average and repetitive behavior, as shown in Figure \ref{fig:trackd}. This suggests that, even though differentiable NMPC is central in our approach, also learning from raw images is essential to provide more flexibility when dealing with unknown or difficult-to-model features that determine a driving style.

\addtolength{\textheight}{-0.2cm}   

\section{CONCLUSIONS}
We presented DriViDOC: a model for Driving from Vision through Differentiable Optimal Control. This model integrates CNN and differentiable NMPC, creating an end-to-end driving model capable of learning the mapping from images to controls. The controls, derived from NMPC, are optimized using a parametric objective function, with dynamic parameters coming from the CNN. Trained on an offline dataset with BC, DriViDOC can successfully imitate different human driving styles, while taking into account lane boundaries, speed and actuators constraints. Moreover, the parameters offer precious insights on how the driving styles are realized.
The model shows improvements with respect to baseline approaches including NMPC and NN. These improvements stem both from incorporating images as inputs and from training with the differentiable NMPC.
Current limitations include the simplicity of both the driving scenarios and the simulated camera images. For more complex situations, additional dynamic agents and traffic rules would need to be incorporated into the NMPC constraints, and more robust approaches, such as control barrier functions, could be integrated into the formulation. Additionally, the sim-to-real gap between simulated and real images should be addressed, potentially through more photorealistic simulations or generative AI solutions. Future work will also involve human-in-the-loop validation of the models.



 \section*{SUPPLEMENTARY MATERIAL}
For transparency, we provide supplementary material at the following link: \href{https://meco-group.github.io/DriViDOC/}{\url{https://meco-group.github.io/DriViDOC/}}. There, we display the closed-loop state simulations and the NMPC parameters for the other drivers, as it is shown for 00 and 01 in Fig.~\ref{fig:simulations}-\ref{fig:params}, along with tables containing metrics for each baseline and for individual drivers, similar to Table \ref{tab:alldrivers}.

\section*{ACKNOWLEDGMENT}
\small
This project has received funding from the Flemish Agency for Innovation and Entrepreneurship (VLAIO) under Baekeland Mandaat No. HBC.2020.2263 (MIMIC) and research project No. HBC.2021.0939 (BECAREFUL) and has been supported by the European Union's Horizon 2020 research and innovation program under the grant agreement No. 953348 (ELO-X). Moreover, we would like to extend our gratitude to Vincent Van Ermengem and all the voluntary drivers for supporting the data collection.

\normalsize
\bibliographystyle{nourl_IEEEtran}
\bibliography{Library}


\end{document}